\documentclass[conference]{IEEEtran}
\IEEEoverridecommandlockouts
\usepackage{cite}
\usepackage{amsmath,amssymb,amsfonts}
\usepackage{algorithmic}
\usepackage{graphicx}
\usepackage{textcomp}
\usepackage{xcolor}
\usepackage{multirow}
\usepackage{graphicx}
\usepackage{makecell}
\usepackage{xspace}
\usepackage{hyperref}
\def\BibTeX{{\rm B\kern-.05em{\sc i\kern-.025em b}\kern-.08em
    T\kern-.1667em\lower.7ex\hbox{E}\kern-.125emX}}
\begin{document}


\makeatletter
\DeclareRobustCommand\onedot{\futurelet\@let@token\@onedot}
\def\@onedot{\ifx\@let@token.\else.\null\fi\xspace}

\def\eg{\emph{e.g}\onedot} \def\Eg{\emph{E.g}\onedot}
\def\ie{\emph{i.e}\onedot} \def\Ie{\emph{I.e}\onedot}
\def\cf{\emph{c.f}\onedot} \def\Cf{\emph{C.f}\onedot}
\def\etc{\emph{etc}\onedot} \def\vs{\emph{vs}\onedot}
\def\wrt{w.r.t\onedot} \def\dof{d.o.f\onedot}
\def\etal{\emph{et al}\onedot}
\makeatother

\title{Integrating Retrospective Framework in Multi-Robot Collaboration }

\author{\IEEEauthorblockN{Jiazhao Liang$^\dagger$\thanks{$\dagger$ indicates equal contribution to this work.} \quad Hao Huang$^\dagger$ \quad Yu Hao \quad Geeta Chandra Raju Bethala}
\IEEEauthorblockN{\quad Congcong Wen \quad John-Ross Rizzo \quad Yi Fang$^*$\thanks{$^*$ indicates the corresponding author.}}
\IEEEauthorblockA{\textit{NYUAD Center for Artificial Intelligence and Robotics (CAIR), Abu Dhabi, UAE} \\
\textit{Embodied AI and Robotics (AIR) Lab, NYU Abu Dhabi, UAE} \\
\textit{New York University Abu Dhabi, UAE} \\
\textit{New York University, US} \\
\texttt{jl9356,hh1811,yh3252,gb2643,cw3437,yfang}@nyu.edu,\texttt{johnrossrizzo}@gmail.com}


}

\maketitle

\begin{abstract}
Recent advancements in Large Language Models (LLMs) have demonstrated substantial capabilities in enhancing communication and coordination in multi-robot systems. However, existing methods often struggle to achieve efficient collaboration and decision-making in dynamic and uncertain environments, which are common in real-world multi-robot scenarios. To address these challenges, we propose a novel \textit{retrospective actor-critic} framework for multi-robot collaboration. This framework integrates two key components: (1) an actor that performs real-time decision-making based on observations and task directives, and (2) a critic that retrospectively evaluates the outcomes to provide feedback for continuous refinement, such that the proposed framework can adapt effectively to dynamic conditions. Extensive experiments conducted in simulated environments validate the effectiveness of our approach, demonstrating significant improvements in task performance and adaptability. This work offers a robust solution to persistent challenges in robotic collaboration.
\end{abstract}

\begin{IEEEkeywords}
Large Language Models, Multi-Robot Collaboration, Actor-Critic
\end{IEEEkeywords}

\section{Introduction}
\label{sec:intro}
Multi-robot collaboration is of a significant interest to the research community due to its potential to enhance efficiency and scalability in complex multi-step tasks across various domains, including manufacturing, search and rescue, and exploration and assistive technologies. 
Despite advances in both areas, key challenges remain in synchronizing task dependencies, managing dynamic environments, and minimizing communication overhead. 
Addressing these challenges is essential to improving the overall effectiveness and robustness of multi-robot collaboration in dynamic and uncertain environments.
Recent research has shown that Large Language Models (LLMs) are being integrated into multi-robot collaboration systems to enhance communication, task planning and coordination. 
For example, in SMART-LLM~\cite{kannan2023smart}, LLMs are used to decompose complex tasks, allocate subtasks to specific robots based on their capabilities, and handle coalition formation for tasks requiring multiple robots to collaborate, \eg, lifting objects. Other studies, such as RoCo~\cite{mandi2024roco}, utilize LLM-driven dialogue systems to optimize high-level task planning and low-level path coordination, improving efficiency in complex multi-robot object manipulation. Swarm-GPT~\cite{jiao2023swarm} proposes a system that integrates LLMs with safe motion planning to automate and ensure collision-free drone swarm choreography. Despite these advances, challenges remain to manage the limits and dynamics given the context of interest. 

To resolve the challenge of dynamically improving decision-making and task execution in multi-robot systems in complex environments, based on the integration of LLMs in multi-robot collaboration~\cite{mandi2024roco}, our paper employs the Reflexion~\cite{shinn2024reflexion,yang2024embodied}, a verbal version of the actor-critic framework \cite{konda1999actor}, to enhance the decision-making process and task execution among multiple robots. This framework divides the responsibilities of each robot between two roles: an \textit{actor}, responsible for generating actions based on current state observations and task instructions, and a \textit{critic}, which evaluates the actions based on their outcomes and then provides feedback. This framework allows for dynamic adjustments and learning, where the actor modifies its actions according to the critic's feedback, leading to more refined and effective actions over time~\cite{yang2024embodied}. The retrospective aspect of the critic enables a deeper analysis of past interactions, identifying and correcting suboptimal or failed actions, thus fostering a continuous learning environment \cite{shinn2024reflexion}. This approach not only improves task performance by adapting to complex environments and challenges but also aligns robotic actions more closely with a strategic objective, reducing operational errors and enhancing overall system robustness. Our key contributions are summarized as follows:

\begin{itemize}
\item  We introduce an innovative approach to multi-robot collaboration by incorporating a retrospective actor-critic framework. This framework enhances decision-making processes by allowing each robot to act (actor) based on the current state observations and task directives while simultaneously receiving evaluative feedback (critic) based on action outcomes.

\item  We conducted extensive simulated experiments, demonstrating that our retrospective actor-critic framework enhances multi-robot collaboration and ensures accurate, robust task execution in challenging scenarios.
\end{itemize}

\begin{figure*}[!htb]
    \centering
    \includegraphics[width=0.99\textwidth]{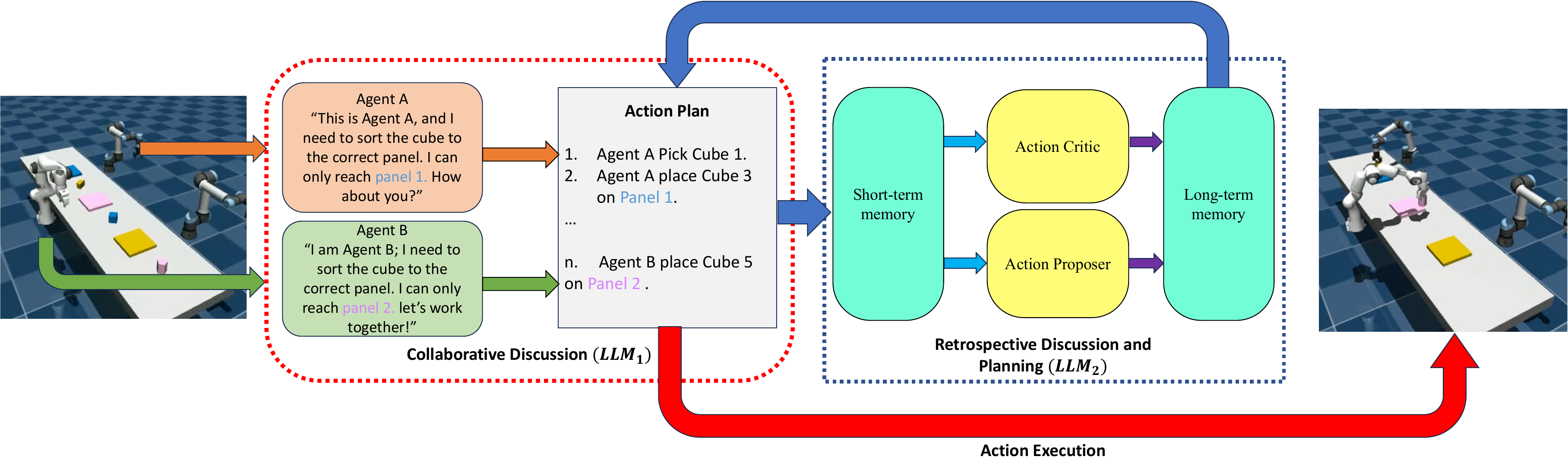}
    \caption{The illustration of a multi-agent collaborative task involving robotic arms (Agent A and Agent B in this case) is coordinated by two large language models, $LLM_1$ and $LLM_2$. In the first stage (collaborative discussion), agents collaboratively discuss an action plan, assigning tasks such as \textit{picking and placing cubes on specific panels}. After validating the plan for inverse kinematics and collision avoidance, it is stored in short-term memory and passed to $LLM_2$ for retrospective discussion and planning. $LLM_2$ contains an action critic that provides a high-level overview for improving performance, while an action proposer offers more detailed suggestions. This feedback is stored in long-term memory. Finally, only two rounds of feedback are selected as the context for $LLM_1$ to produce an improved action plan, which is then executed in the environment.}
    \label{fig:pipeline}
\end{figure*}

\section{Related Work}
\label{sec:related}

\noindent \textbf{Multi-robot Collaboration.}
Research on multi-robot manipulation is extensive, with early work concentrating on low-level planning, such as devising collision-free motion trajectories \cite{xie2023language}, where sampling-based methods have been particularly prominent \cite{karaman2011sampling}. Recent advancements have also introduced learning-based approaches as viable alternatives \cite{zhang2022multi}. Additionally, some works focus on multi-robot collaboration in dynamic and complex control scenarios \cite{zeng2022socratic}. Moreover, the development of embodied AI simulators has facilitated the emergence of embodied collaboration tasks, such as collaborative navigation \cite{liu2022multi} and furniture rearrangement \cite{jain2020cordial}. 

\noindent \textbf{LLM in Robotics.}
Recent robotics research integrates large language models (LLMs) to enhance natural language understanding~\cite{khurana2023natural} and navigation~\cite{wen2025zero,wen2024secure}. This integration allows vision-language models~\cite{zhang2021vinvl} to improve visual question answering~\cite{lin2023medical} and support multi-modal planning~\cite{shao2023prompting} through combined textual, visual, and auditory input. Methods such as Inner Monologue~\cite{huang2022inner} and ManipLLM~\cite{li2024manipllm} demonstrate how LLM-driven reasoning enhances interaction and manipulation tasks. Most recently, RoCo~\cite{mandi2024roco} explores multi-robot collaboration, integrating high-level language-based guidance with low-level path planning. Yet, LLMs still face challenges in data-intensive trial-and-error learning~\cite{shinn2024reflexion}. To address this issue, we propose a retrospective actor-critic framework that enables multi-robot teams to learn effectively from previous failures and refine future decisions.

\section{Method}
\label{sec:method}

\subsection{Overview}
\label{subsec:pipeline}
The overview of our multi-robot system is illustrated in Figure~\ref{fig:pipeline}. Following~\cite{mandi2024roco}, our system integrates collaborative discussion and action execution. In addition, it incorporates \textit{retrospective discussion and planning}~\cite{yang2024embodied} for action planning, which is proposed in this work. The system involves $N$ agents\footnote{Here, for a concise description, we set the number of agents $N$ to 2, but it can be easily extended to include more agents.}, with each assigned a specific role. A task description $T$, containing the task goal, is provided to the robots. At time $t$, each robot has its current observation denoted as $o_t^n$.

Each robot, assigned a specific role, participates in discussions simulated by $LLM_1$, which manages communication between agents. In each dialogue turn $k$, robot $n$ generates a message $m_k^n = LLM_1(p_k^n)$, where $p$ is the prompt containing the observation $o_t^n$ and task $T$. After the discussion, proposed actions are validated by checking subgoal feasibility, collision risk, and inverse kinematics issues. Once validated, this information is stored in short-term memory. Then, these plans are sent to $LLM_2$, an actor-critic structure~\cite{yang2024embodied}, for retrospective analysis, which evaluates actions, identifies improvements, and gives feedback. The long-term memory retains only the most recent two retrospectives, serving as context for future discussions. This iterative process refines the action plans, leading to continuous task improvement and successful execution.

\subsection{Retrospective Analysis}
\label{subsec:retrosp}
The core of our retrospective analysis framework is the \textit{actor-critic} structure~\cite{yang2024embodied}, which leverages LLM-based reflective capabilities~\cite{shinn2024reflexion}. In this setup, $LLM_1$ generates collaborative conversations among robots to agree on subgoals and actions. A validation step then assesses the feasibility of these actions, ensuring subgoals are achievable and IK solutions are collision-free. Any issues, such as collisions or IK failures, are stored in short-term memory along with the conversations, forming part of the prompt for $LLM_2$.

With this accumulated information, the actor-critic mechanism is activated. The combined data from the validation results and the initial conversation are fed into $LLM_2$. Acting as an action critic, $LLM_2$ scrutinizes all previous information to identify any issues with the proposed actions. It generates a detailed critique that emphasizes the feedback, highlighting specific problems such as potential collisions, unreachable subgoals, or IK failures. For example, $LLM_2$ will provide the following criticism: ``The path for \textit{Agent A} is too low; hence, it collided with the other agent; please increase the height next time.'' This critique is incorporated into the prompt to ensure that the robots are fully aware of the aspects of the plan that require attention.

Following the critique phase, $LLM_2$ is now in charge of the action proposer. It provides insights on how to improve the current action plan by offering alternative viewpoints and suggesting adjustments to address the identified obstacles. This may involve proposing new subgoals, reordering action sequences, or reallocating tasks among the robots to enhance performance. For instance, as seen earlier, the critic noted how an agent caused a collision. Then, the action proposer analyzes this criticism and proposes a more specific action: ``Agent A should increase the height by 0.5 to avoid a collision.'' This highlights the complementary nature of the two roles: the critic identifies the problem, and the action proposer suggests precise modifications to refine future actions. By considering the collective capabilities and limitations of all robots involved, $LLM_2$ delivers a more comprehensive perspective that transcends individual viewpoints.

The output from $LLM_2$ thus consists of two integral parts: the critique of the proposed actions and the potential improvement to the action plan. The critique pinpoints flaws and potential risks in the initial plan, ensuring that critical issues are not overlooked. The improved action plan offers practical recommendations for adjustments, enabling the robots to refine their strategies effectively.

All interactions—initial conversations facilitated by $LLM_1$, validation feedback, and environmental feedback—are first stored in short-term memory. With information from short-term memory, $LLM_2$ promptly analyzes actions, identifies issues, and suggests improvements. The critiques and advice from $LLM_2$ are then stored in long-term memory, which retains only the most recent two rounds of interactions, including initial conversations, validation feedback, and critique suggestions. The information in long-term memory is then fed back into $LLM_1$ as part of the prompt for subsequent planning. Mathematically, we use $S$ to denote short-term memory, and at time $t$, $S_{t-1}$ and $S_t$ are combined into long-term memory $M_{i+1}$. $M_{i+1}$ is then used as context to enhance future prompts: 
\begin{equation}
p_{k+1}^n = \texttt{ConstructPrompt}(o_{t+1}^n, T, M_{t+1})
\end{equation}
By incorporating recent experiences, the agents become more aware of past mistakes while maintaining efficiency in the planning process. 

\subsection{Prompt Design}
\label{subsec:prompt}
\noindent \textbf{Task prompt design.}
We adhere to the design of the original RoCo bench~\cite{mandi2024roco} for our task prompts, which contain the following components: task context, round history, agent capability, communication guidelines, observation, and feedback. The \emph{task context} provides a high-level overview of the task's objective; the \emph{round history} records previous conversations and actions; \emph{agent capability} describes the agent's available skills and constraints; \emph{communication guidelines} specify how agents should interact and the expected output format; the \emph{observation} details the task and the current object the robot is holding; and \emph{feedback} ensures validation so that the robot's inverse kinematics computations do not fail and robot paths do not intersect, spurring collisions.


\begin{table}[htbp]
\centering
\scriptsize
\caption{Evaluation results on RoCoBench~\cite{mandi2024roco}. We report averaged success rates ($\uparrow$) over 15 runs per task, the average number of steps in successful runs ($\downarrow$), and the average number of re-plan attempts ($\downarrow$) used across all runs.}
\label{table:sim_vertical}
\begin{tabular}{|c|c|c|}
\hline
\textbf{Approach} & \textbf{RoCo} \cite{mandi2024roco} & \textbf{Ours} \\ 
\hline
Metrics & \makecell{Success (\%), Steps, Replan} & \makecell{Success (\%), Steps, Replan} \\
\hline
Arrange Cabinet & 
\makecell{0.27\%±0.12, 8.9, 3.1} & 
\makecell{\textbf{0.40\%}±0.13, \textbf{8.1}, 3.1} \\
\hline
Sweep Floor & 
\makecell{0.07\%±0.07, 9.9, 1.1} & 
\makecell{\textbf{0.20\%}±0.10, \textbf{9.7}, 1.1} \\
\hline
Make Sandwich & 
\makecell{0.13\%±0.09, \textbf{8.3}, 2.3} & 
\makecell{\textbf{0.20\%}±0.10, 9.7, \textbf{2.2}} \\
\hline
Sort Cubes & 
\makecell{0.33\%±0.12, 8.9, 3.1} & 
\makecell{\textbf{0.40\%}±0.13, \textbf{8.1}, \textbf{3.0}} \\
\hline
Move Rope & 
\makecell{0.27\%±0.12, 5.6, 4.2} & 
\makecell{\textbf{0.33\%}±0.12, \textbf{4.6}, \textbf{3.8}} \\
\hline
\end{tabular}
\end{table}

\noindent \textbf{Retrospection prompt design.}
To enhance the decision-making capabilities of our agents, we implement an \textit{actor-critic} prompt design scheme, which encourages each agent to focus on their own current plan while actively incorporating feedback from the environment to iteratively improve their actions. The actor-critic scheme operates through two interconnected components:
\begin{enumerate} 
\item \textbf{Critic prompt}: Agents are also prompted to critically evaluate the proposed actions—either their own or those of other agents. This involves analyzing feedback from the environment and identifying rooms where the actions could be enhanced or adjusted to better meet task requirements. An example prompt for this is ``based on the feedback and the conversation among agents, please critique the plan agreed upon and what can possibly be done to improve performance.''
\item \textbf{Actor prompt}: Agents are prompted to propose action revisions based on the current state and task objectives. They are guided to consider how their actions contribute to the overall plan and to focus on achieving specific goals efficiently. An example prompt is ``please provide detailed suggestions on how each agent should modify the current plan.'' 
\end{enumerate}
By integrating these components into the prompt design, agents engage in a continuous loop of action and reflection. They not only execute actions but also assess the effectiveness of those actions, learning from previous successful and failed cases for the next round decision-making.



\section{Experiments}
\label{seb:exp}


To evaluate the effectiveness of our proposed actor-critic framework for multi-robot collaboration, we conducted extensive experiments using the RoCoBench simulation environment \cite{mandi2024roco} and followed the same experimental settings. 
%
RoCoBench comprises six diverse robot collaboration manipulation tasks involving commonly seen objects. We selected five tasks that are specifically designed to test the effectiveness of the robot collaboration and task planning:

 \begin{itemize}
     \item \textbf{Arrange Cabinet:} The goal of this task is to pick a cup from the cabinet and place the cup on the table. Two robots need to work together to open the door, and the third robot is responsible for placing the cup.
     \item \textbf{Sweep Floor:} This task contains three objects spread on the table, with one robot holding the dustpan and the other holding the broom. The object here is to use the broom to sweep all objects on the table into the dustpan and throw them away.
     \item \textbf{Make Sandwich:} In this task, some ingredients are spread on the table. Robots need to collaboratively pick up ingredients and stack them in the correct order to make a sandwich.
     \item \textbf{Sort Cube:} In this task, there are three cubes with different colors, and three robots need to collaborate to pick and place the cube on the pad with the same color.
     \item \textbf{Move Rope:} This experiment requires two robots to pick up the rope lying on the table and place it on a tray at a designated orientation.
 \end{itemize}

For the model selection, $LLM_1$ is based on Llama3.1-70b model \cite{dubey2024llama}for generating actions, and the actor-critic in $LLM_2$ was implemented based on Llama3.1-8b model. We conducted each task 20 times, and the performance metrics can be found in Table \ref{table:sim_vertical}. For a fair comparison, we used the open source code\footnote{\url{https://github.com/MandiZhao/robot-collab}.} released by RoCo~\cite{mandi2024roco} but replaced GPT-4~\cite{achiam2023gpt} with Llama3.1-70b to reproduce their results. We used four NVIDIA A100 GPUs (80 GB each) to run both models in parallel, ensuring adequate memory for real-time inference and efficient parallel operations.

\begin{figure*}[htbp]
\centering
\includegraphics[width=.9\textwidth]{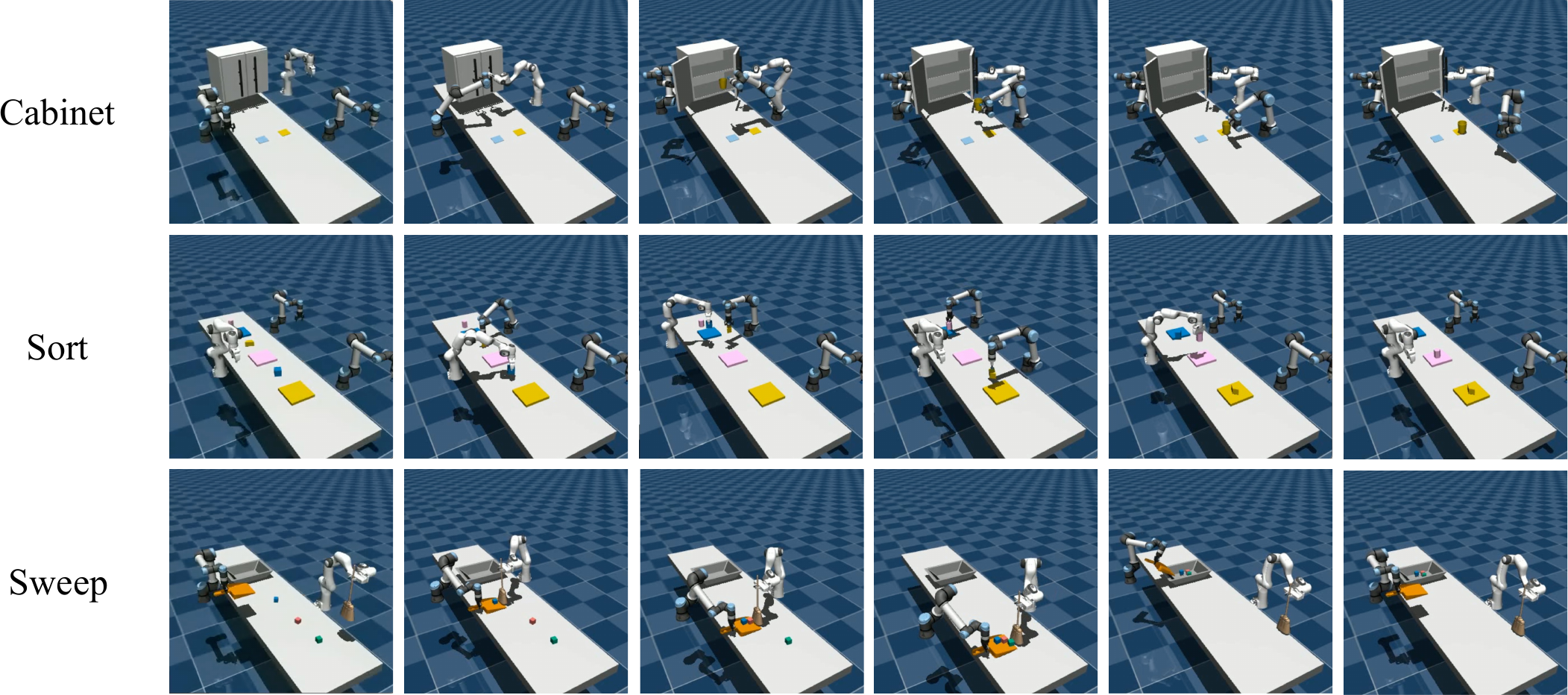}
\caption{Simulation qualitative results. Here, we demonstrate three tasks completed by multiple agents. Each row represents the steps performed by multiple robots in cooperation for a given task.}
\label{fig:sim}
\end{figure*}

Our approach outperforms the RoCo baseline across multiple tasks. On arrange cabinet, the success rate increased from RoCo’s 27\% to 40\%, and on sort cubes, it rose from 33\% to 40\%. Similarly, for the move rope task, our success rate of 33\% surpasses RoCo's 27\%, underscoring the effectiveness of incorporating environmental feedback into our approach.
Moreover, our method also enhances efficiency by balancing fewer steps with reduced re-planning instances. For example, in the arrange cabinet task, the average number of steps decreased from RoCo's 8.9 to 8.1 with our method. In the move rope task, the improved success rates were accompanied by fewer re-planning events, highlighting our method's superior task efficiency and adaptability to feedback.
 Figure~\ref{fig:sim} illustrates our approach’s advantages and improvements over the baseline.

%

%


\section{Ablation Study}
We conducted two experiments to evaluate our design. The first experiment tested how restricting background information to a single round affects the generation of actions, and the second experiment evaluated how the choice of model affects our results. We replaced the Llama 3.1 70B model with NVIDIA’s Nemotron-70B \cite{wang2024helpsteer2preferencecomplementingratingspreferences}.

\subsection{Context Length}
In the original design, the model processes two rounds of background history, allowing it to capture rich context and provide more effective critiques. Here, we limit it to a single round to examine the impact of reduced context on performance. Results are shown in Table \ref{table:ablation_context_vertical}.

%

\begin{table}[htbp]
\centering
\scriptsize
\caption{Evaluation on Context Length}
\label{table:ablation_context_vertical}
\begin{tabular}{|c|c|c|}
\hline
\textbf{Approach} & \textbf{Single Length Memory} & \textbf{Ours} \\ 
\hline
Metrics & \makecell{Success (\%), Steps, Replan} & \makecell{Success (\%), Steps, Replan} \\
\hline
Arrange Cabinet & 
\makecell{0.27±0.12, \textbf{8.8}, \textbf{2.9}} & 
\makecell{\textbf{0.40±0.13}, \textbf{8.1}, 3.1} \\
\hline
Sweep Floor & 
\makecell{0.20±0.10, \textbf{9.2}, 1.4} & 
\makecell{0.20±0.10, 9.7, \textbf{1.1}} \\
\hline
Make Sandwich & 
\makecell{0.13±0.09, \textbf{9.2}, \textbf{2.1}} & 
\makecell{\textbf{0.20±0.10}, 9.7, 2.2} \\
\hline
Sort Cubes & 
\makecell{0.20±0.10, 8.1, 3.1} & 
\makecell{\textbf{0.40±0.13}, 8.1, \textbf{3.0}} \\
\hline
Move Rope & 
\makecell{0.20±0.10, 5.6, \textbf{3.7}} & 
\makecell{\textbf{0.33±0.12}, \textbf{4.6}, 3.8} \\
\hline
\end{tabular}
\end{table}

As shown in the table, reducing the amount of contextual information negatively affects the model's performance. Specifically, the success rate for the sort task’s success rate fell by 20\%, the cabinet task by 27\%, and the rope task by 20\%. Although the sweep task success rate remained stable, the number of replans increased, suggesting reduced stability.



\subsection{Model Alteration}
In the second experiment, We replaced the Llama 3.1 70B model \cite{dubey2024llama} with NVIDIA's Nemotron LLM 70B model \cite{wang2024helpsteer2preferencecomplementingratingspreferences}. This change aimed to assess the influence of different model architectures and training data on action generation and critiques within our framework. The results, presented in Table \ref{table:ablation_model_vertical}, show notable variations across tasks. The success rate for the sandwich task increased by 13\%, while the sort and sweep tasks experienced decreases to 13\% and 7\%, respectively.


\begin{table}[htbp]
\centering
\scriptsize
\caption{Evaluation on Model Alteration}
\label{table:ablation_model_vertical}
\begin{tabular}{|c|c|c|}
\hline
\textbf{Approach} & \textbf{Single Length Memory} & \textbf{Ours} \\ 
\hline
Metrics & \makecell{Success (\%), Steps, Replan} & \makecell{Success (\%), Steps, Replan} \\
\hline
Arrange Cabinet & 
\makecell{0.33±0.12, \textbf{7.1}, 3.6} & 
\makecell{\textbf{0.40±0.13}, 8.1, \textbf{3.1}} \\
\hline
Sweep Floor & 
\makecell{0.07±0.07, \textbf{8.2}, 1.2} & 
\makecell{\textbf{0.20±0.10}, 9.7, \textbf{1.1}} \\
\hline
Make Sandwich & 
\makecell{\textbf{0.33±0.12}, \textbf{9.4}, \textbf{2.1}} & 
\makecell{0.20±0.10, 9.7, 2.2} \\
\hline
Sort Cubes & 
\makecell{0.13±0.09, 8.2, 4.0} & 
\makecell{\textbf{0.40±0.13}, \textbf{8.1}, \textbf{3.0}} \\
\hline
Move Rope & 
\makecell{0.20±0.10, 5.8, 4.4} & 
\makecell{\textbf{0.33±0.12}, \textbf{4.6}, \textbf{3.8}} \\
\hline
\end{tabular}
\end{table}




These results underscore the importance of model selection. While Nemotron 70B boosted performance in the sandwich task, it failed to improve other collaboration tasks. This suggests that different models may excel at different tasks, underscoring the variability in model-task compatibility. Thus, choosing a model that aligns with the application’s specific needs is crucial.
\section{Conclusion}
\label{sec:con}
Despite the advances brought about by LLMs, multi-robot systems continue to face significant challenges related to efficient decision-making and dynamic adaptation within complex environments. We introduced a retrospective actor-critic framework that harnesses LLMs to improve both immediate and long-term decision-making in robotics. Through extensive experiments, our results demonstrate that the retrospective actor-critic framework significantly boosts effectiveness, adaptability, and robustness of multi-robot systems. However, using an LLM without fine-tuning occasionally produced “hallucinated” outputs, leading to suboptimal planning and resource allocation.

\bibliographystyle{IEEEtran}
\bibliography{egbib}

\end{document}